\documentclass[11pt]{article}
\usepackage[preprint]{acl}

\usepackage{times}
\usepackage{latexsym}

\usepackage[T1]{fontenc}

\usepackage[utf8]{inputenc}

\usepackage{microtype}

\usepackage{inconsolata}

\usepackage{graphicx}

\usepackage{amsmath}
\usepackage{amssymb}
\usepackage{booktabs}
\usepackage{multirow}
\usepackage[table]{xcolor}
\usepackage{arydshln}
\usepackage[linesnumbered,ruled,vlined]{algorithm2e}
\usepackage{listings}
\usepackage{tcolorbox}
\usepackage{courier}
\usepackage{hyperref}
\usepackage{xcolor}

\title{Step Potential Advantage Estimation: Harnessing Intermediate Confidence and Correctness for Efficient Mathematical Reasoning}

\author{
  \textbf{Fei Wu}\textsuperscript{1}\thanks{\ \ Equal contribution}, \quad 
  \textbf{Zhenrong Zhang}\textsuperscript{1,2}\footnotemark[1], \quad 
  \textbf{Qikai Chang}\textsuperscript{1}, \quad 
  \textbf{Jianshu Zhang}\textsuperscript{2}, \quad 
  \\
  \textbf{Quan Liu}\textsuperscript{2}, \quad 
  \textbf{Jun Du}\textsuperscript{1}\thanks{\ \ Corresponding author: Jun Du (jundu@ustc.edu.cn).}, \\
  \textsuperscript{1}University of Science and Technology of China \\
  \textsuperscript{2}iFLYTEK Research \\
}

\begin{document}
\maketitle

\begin{abstract}
Reinforcement Learning with Verifiable Rewards (RLVR) elicits long chain-of-thought reasoning in large language models (LLMs), but outcome-based rewards lead to coarse-grained advantage estimation. While existing approaches improve RLVR via token-level entropy or sequence-level length control, they lack a semantically grounded, step-level measure of reasoning progress. As a result, LLMs fail to distinguish necessary deduction from redundant verification: they may continue checking after reaching a correct solution and, in extreme cases, overturn a correct trajectory into an incorrect final answer. To remedy the lack of process supervision, we introduce a training-free probing mechanism that extracts intermediate confidence and correctness and combines them into a \emph{Step Potential} signal that explicitly estimates the reasoning state at each step. Building on this signal, we propose \emph{Step Potential Advantage Estimation} (SPAE), a fine-grained credit assignment method that amplifies potential gains, penalizes potential drops, and applies penalty after potential saturates to encourage timely termination. Experiments across multiple benchmarks show SPAE consistently improves accuracy while substantially reducing response length, outperforming strong RL baselines and recent efficient reasoning and token-level advantage estimation methods. The code is available at \url{https://github.com/cii030/SPAE-RL}.
\end{abstract}

\begin{figure}[t]
    \centering
    \includegraphics[width=\linewidth]{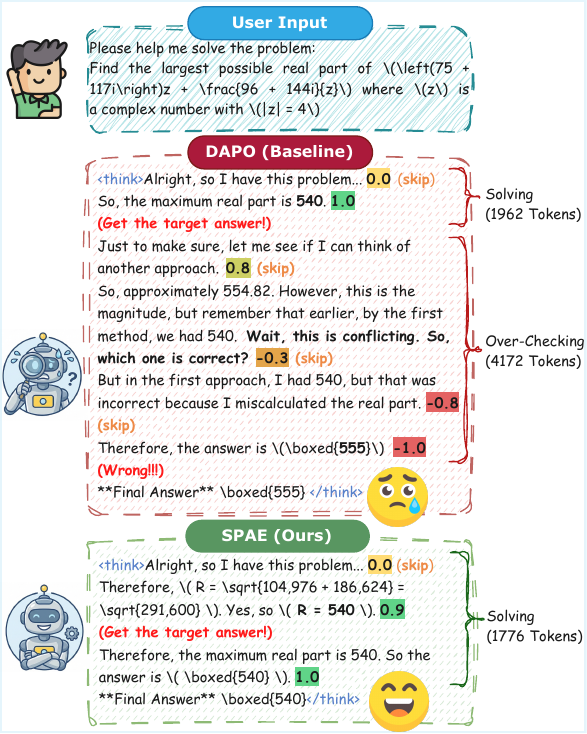}
    \caption{Given the same user query, the baseline RLVR model DAPO produces a Right-to-Wrong failure due to over-checking, whereas SPAE terminates confidently after reaching the correct solution. The {\setlength{\fboxsep}{1pt}\colorbox{blue!10}{value}} with colored background appended after each step indicates the corresponding Step Potential.}
    \label{fig:teaser}
\end{figure}

\section{Introduction}

Reinforcement Learning with Verifiable Rewards (RLVR) has become a central paradigm for eliciting long chain-of-thought (CoT) reasoning in large language models (LLMs) \citep{wei2022chain, openai_learning_2024, guo2025deepseek, li2025system}. By optimizing outcome-level correctness, RLVR aligns model behavior with verifiable task success and delivers substantial gains on mathematics, logic, and coding benchmarks. However, RLVR provides sparse supervision since reward arrives only after the full generation is complete \citep{sun2025ktae}. This outcome-only feedback makes credit assignment ambiguous: the policy cannot reliably identify which parts of a trajectory are essential to reaching the solution and which are merely incidental. In practice, this ambiguity often manifests as unnecessarily long and circuitous reasoning.

Recent work mitigates this issue with finer-grained heuristics. One dominant line uses token entropy as a proxy for importance, positing that high-entropy tokens correspond to exploration \citep{cheng2025reasoningexplorationentropyperspective, chen2025seedgrposemanticentropyenhanced, wang20258020rulehighentropyminority}. Another line regularizes verbosity by rewarding correctness while penalizing length, encouraging early stopping \citep{zhang-etal-2025-adaptthink, shen-etal-2025-dast, cheng2025optimizing}. These approaches largely operate at the token or sequence level and remain agnostic to semantic progress. Crucially, both categories lack a step-level estimate of reasoning progress that can distinguish necessary deduction from redundant verification.

To bridge this gap, we introduce a training-free probing mechanism that makes step-wise progress observable. After each reasoning step, we prompt the model to produce a tentative answer and extract two intermediate signals: confidence and correctness. We then combine them into \emph{Step Potential} that estimates current reasoning state: high potential indicates justified confidence, while low potential reflects unreliable reasoning.

\begin{figure}[t]
    \centering
    \includegraphics[width=\linewidth]{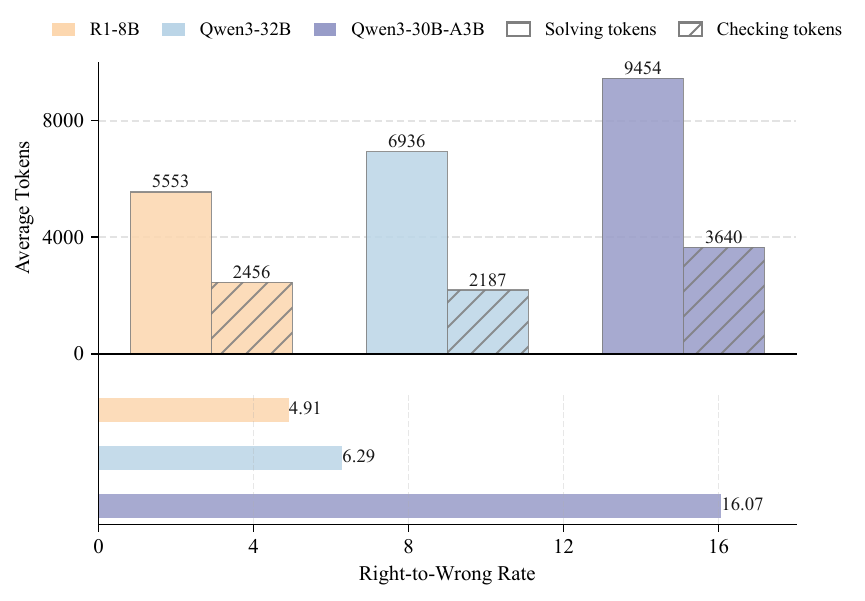}
    \caption{Quantifying Over-Checking on AIME2024 (averaged over 16 samples). Top: average solving and checking tokens on correct responses. Bottom: Right-to-Wrong (R2W) rate on incorrect responses.}
    \label{fig:overchecking}
\end{figure}

Step Potential enables a direct diagnosis of a failure mode we term \emph{Over-Checking}. Once the reasoning model has solved the problem (Step Potential saturates), it often continues to generate redundant post-solution verification. More importantly, prolonged verification increases the risk of a \emph{Right-to-Wrong (R2W) Failure}: after reaching a correct solution, the model continues checking and eventually revises its answer to a wrong one (Figure~\ref{fig:teaser}). Figure~\ref{fig:overchecking} quantifies this phenomenon by separating tokens into solving and checking phases, and reports the R2W rate on incorrect trajectories. Notably, even the latest strong model Qwen3-32B \citep{yang2025qwen3technicalreport} produces substantial redundant checking tokens and still exhibits R2W failures.

Building on Step Potential, we propose \emph{Step Potential Advantage Estimation} (SPAE), an RL method that directly incorporates step-level potential into policy optimization. SPAE improves redundancy control and credit assignment by (1) applying a potential saturation penalty to encourage timely termination once the model has reached a correct solution, and (2) amplifying advantages for pivotal transitions that induce large potential increases, penalizing steps that decrease potential

We evaluate SPAE on challenging mathematical benchmarks and out-of-distribution tasks with multiple reasoning models (Qwen and Llama families). SPAE consistently outperforms strong RL baselines and closely related efficient reasoning and token-level advantage estimation methods. Beyond improving accuracy, SPAE effectively prunes redundant post-solution verification, reducing inference cost without sacrificing performance. For example, on AIME2024, AIME2025, and GPQA, SPAE reduces the average response length of DeepSeek-R1-Distill-Qwen-7B by 25.1\%, 25.3\%, and 24.4\%, while improving accuracy by 6.7\%, 3.3\%, and 1.1\%, respectively.

Our contributions are summarized as follows:
\begin{itemize}
    \item We introduce a training-free probing mechanism and the Step Potential metric, and formalize Over-Checking as a pathological behavior that degrades efficiency and can trigger Right-to-Wrong failures.
    \item We propose SPAE, a step-aware RL credit assignment method that encourages large gains in Step Potential, penalizes declines, and suppresses redundant post-solution checking.
    \item Extensive experiments show that SPAE simultaneously improves accuracy and reduces inference length across models and benchmarks.
\end{itemize}

\begin{figure*}[t]
  \includegraphics[width=\linewidth]{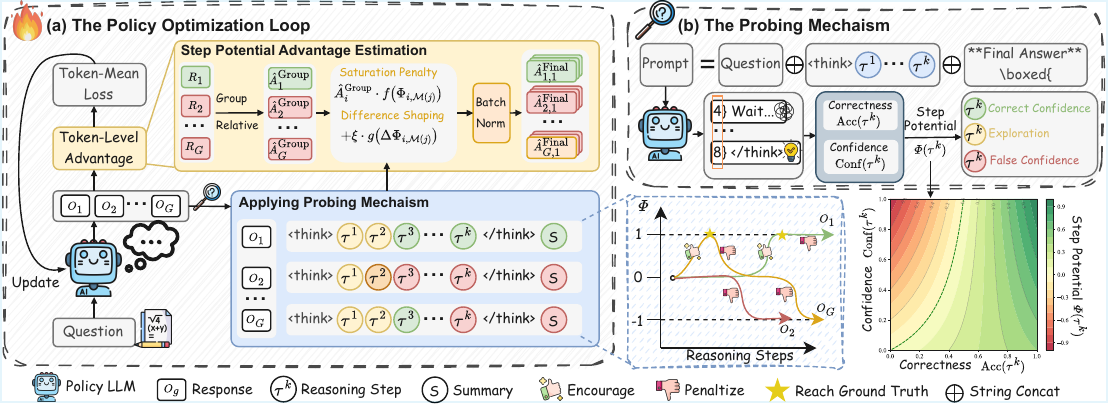}
  \caption{\textbf{Overview of our proposed method.} (a) SPAE integrates Step Potential from the probing mechanism into the RLVR optimization loop by combining group-relative outcome advantages with step-aware credit assignment after each rollout. (b) Our training-free probing mechanism estimates Step Potential by inserting a prompt after each reasoning step to compute confidence and correctness from the model’s induced answers; the bottom panel visualizes the resulting Step Potential values as a 2D contour over the (Acc, Conf) space.}

  \label{fig:overview}
  \vspace{-10pt}
\end{figure*}

\section{Preliminary}

\subsection{Problem Formalization}

We optimize an LLM policy $\pi_\theta$ for mathematical reasoning. Given a query $q \sim \mathcal{D}$ with ground-truth answer $y^*$, the model generates an output $o \sim \pi_\theta(\cdot \mid q)$. In long-CoT models such as DeepSeek-R1 \citep{guo2025deepseek} and Qwen3 \citep{yang2025qwen3technicalreport}, $o$ comprises a reasoning trajectory $\tau$ (typically wrapped by \texttt{<think>} and \texttt{</think>}) and a final summary $s$, i.e., $o=[\tau; s]$. We segment $\tau=[\tau^{1},\dots,\tau^{K}]$ into $K$ discrete reasoning steps, where each step is a contiguous token span separated by explicit delimiters (e.g., ``.\textbackslash n\textbackslash n''). These steps are the basic units of analysis in this work. RLVR maximizes the expected task-level reward:
\begin{equation}
\mathcal{J}(\theta)=\mathbb{E}_{q \sim \mathcal{D},\, o \sim \pi_\theta(\cdot|q)}\!\left[ R(o,y^*) \right],
\end{equation}
where $R(o,y^*)\in\{0,1\}$ is a binary reward that checks whether the answer extracted from $s$ matches $y^*$.

\subsection{Reinforcement Learning with Verifiable Reward}

\paragraph{Group Relative Policy Optimization (GRPO).}
To avoid value-network overhead in Proximal Policy Optimization \citep{schulman2017ppo}, GRPO \citep{shao2024deepseekmath} estimates baselines from group statistics. For each $q \sim \mathcal{D}$, GRPO samples $G$ responses $\{o_i\}_{i=1}^G \sim \pi_{\theta_{\text{old}}}(\cdot|q)$, computes rewards $\{R(o_i,y^*)\}_{i=1}^G$, and maximizes:
\begin{equation}
\begin{aligned}
    \mathcal{J}_{\text{GRPO}}(\theta) &= \mathbb{E}_{q \sim \mathcal{D},\, \{o_i\}_{i=1}^G \sim \pi_{\theta_{\text{old}}}(\cdot|q)} \bigg[ \\
    &\quad \frac{1}{G} \sum_{i=1}^G \frac{1}{|o_i|} \sum_{j=1}^{|o_i|} \min \Big( r_{i,j}(\theta)\,\hat{A}_{i,j}, \\
    &\quad \operatorname{clip}\!\left(r_{i,j}(\theta), 1-\epsilon, 1+\epsilon\right)\hat{A}_{i,j} \Big) \bigg],
\end{aligned}
\end{equation}
where $\epsilon$ is the clipping threshold, and
\begin{equation}
\begin{aligned}
r_{i,j}(\theta) &= \frac{\pi_\theta(o_{i,j} \mid q, o_{i,<j})}{\pi_{\theta_{\text{old}}}(o_{i,j} \mid q, o_{i,<j})},\\
\hat{A}_{i,j}=\hat{A}_i &= \frac{R(o_i,y^*)-\operatorname{mean}(\{R(o_k,y^*)\}_{k=1}^G)}{\operatorname{std}(\{R(o_k,y^*)\}_{k=1}^G)}.
\end{aligned}
\end{equation}
GRPO assigns a single scalar advantage $\hat{A}_i$ to all tokens in $o_i$, eliminating the critic but yielding coarse credit assignment over the full trajectory.

\paragraph{Decouple Clip and Dynamic Sampling Policy Optimization (DAPO).}
DAPO \citep{yu2025dapo} replaces sample-level averaging with global token-level normalization to better handle varying response lengths. DAPO further improves exploration via Clip-Higher with decoupled bounds $(\epsilon_{\text{low}},\epsilon_{\text{high}})$ and applies Dynamic Sampling, updating query groups with non-zero reward variance.

\paragraph{Reinforce++-Baseline (RF-B).}
To reduce instability under small group sizes $G$, RF-B \citep{hu2025reinforce} standardizes advantages over the full batch $\mathcal{B}$ rather than within each group, alleviating the high-variance local statistics in GRPO/DAPO. The benefits of such global normalization are supported by prior work \citep{liu2025itrickstrapsdeep, mai2025agentrlscalinglaw}.

\paragraph{Our Backbone.} By synthesizing DAPO (exploration strategies) and RF-B (global batch advantage normalization), we establish a robust RLVR baseline. This sets the stage for our proposed fine-grained, step-aware credit assignment mechanism.

\section{Methodology}

As illustrated in Figure~\ref{fig:overview}, we propose SPAE to bridge the gap between sparse rewards and step-level reasoning quality. Our framework proceeds in three stages: (1) extracting intermediate confidence and correctness signals via a training-free probe; (2) synthesizing these signals into Step Potential to distinguish effective solving from redundant verification; and (3) optimizing the policy with potential-aware advantages that amplify pivotal deductions while penalizing Over-Checking.

\subsection{The Probing Mechanism}

To mitigate reward sparsity in long CoT reasoning, we introduce a training-free probing mechanism. This functions as a semantic sensor to convert opaque hidden states into explicit progress signals without requiring auxiliary value networks.

We operate directly at the step level. At the boundary of each step $\tau_i^{k}$ in a response $o_i$ (e.g. ``.\textbackslash n\textbackslash n''), we insert a trigger prompt $p_{\text{probe}}$ ``**Final Answer** \textbackslash n\textbackslash\textbackslash boxed\{'' to induce a tentative conclusion. The probing context is defined as:
\begin{equation}
h_{i,k} = (q, o_{i,\le \tau_i^{k}}, p_{\text{probe}}),
\end{equation}
where $o_{i,\le \tau_i^{k}}$ denotes the response prefix ending at step $k$. Conditioned on this context, the model generates $N$ short continuations $\{Y_1, \dots, Y_N\}$, from which we extract two dense signals characterizing the quality of the reasoning step $\tau_i^{k}$. Each probe continuation is capped at 10 tokens, which is sufficient to complete the induced final answer, and uses the same generation settings as the main reasoning process (temperature, top-$k$, and top-$p$).

\paragraph{Confidence.}
The confidence $\mathrm{Conf}(\tau_i^{k})$ measures how certain the model is about its current conclusion: high entropy indicates uncertainty and fluctuating next-token preferences, while low entropy suggests a more concentrated, converged distribution. For each probe continuation $Y_n=(y_{n,1},\dots,y_{n,L_n})$ under context $h_{i,k}$, let $p_{\theta}(\cdot \mid h_{i,k}, y_{n,<l})$ denote the next-token distribution at position $l$. We compute the token-level entropy:
\begin{equation}
\begin{aligned}
H_{n,l}
\;=\;
- \sum_{v \in \mathcal{V}}
p_{\theta}\!\left(v \mid h_{i,k}, y_{n,<l}\right) \\
\qquad\cdot \log p_{\theta}\!\left(v \mid h_{i,k}, y_{n,<l}\right),
\end{aligned}
\end{equation}
and convert the length-normalized entropy into a bounded confidence score in $[0,1]$ via $\exp(\cdot)$, averaged over $N$ probe samples:
\begin{equation}
\mathrm{Conf}(\tau_i^{k})
\;=\;
\frac{1}{N} \sum_{n=1}^{N}
\exp\!\left(
-\frac{1}{L_n} \sum_{l=1}^{L_n} H_{n,l}
\right),
\end{equation}

\paragraph{Correctness.} The correctness metric (represented by accuracy $\mathrm{Acc}(\tau_i^{k})$) quantifies how compatible the current reasoning prefix is with the ground-truth answer $y^*$. Instead of binary exact matching, we compute a continuous score by force-feeding the ground-truth tokens and averaging their conditional probabilities:
\begin{equation}
\mathrm{Acc}(\tau_i^{k}) = \frac{1}{N} \sum_{n=1}^{N} \frac{1}{|y^*|} \sum_{m=1}^{|y^*|} \pi_\theta\!\left( y_m^* \mid h_{i,k},\, y_{<m}^* \right),
\end{equation}
This formulation yields a dense signal in $[0,1]$, reflecting how likely the model is to produce the ground-truth final answer conditioned on the reasoning prefix up to step $k$.

Importantly, our probing mechanism is training-free and non-intrusive: it introduces no auxiliary parameters, does not backpropagate gradients through probe generations, and does not alter the policy optimization objective. Ground-truth answers are used solely to extract diagnostic signals.

\subsection{Quantifying Reasoning Steps via Step Potential}

Harnessing the fine-grained confidence and correctness signals extracted by the probe, we introduce a unified scalar metric \textit{Step Potential} to quantify the quality of an intermediate reasoning step, denoted as $\Phi(\tau_i^{k})$. Inspired by classical potential-based reward shaping in RL \citep{Ng1999PolicyIU}, Step Potential maps each reasoning step into a bounded range $[-1, 1]$, providing a consistent measure of progress throughout a long reasoning trajectory:
\begin{equation}
\begin{aligned}
\Phi(\tau_i^{k}) = {} &
1.5 \cdot \mathrm{Acc}(\tau_i^{k}) \cdot \mathrm{Conf}(\tau_i^{k}) \\
& {} + 0.5 \cdot \mathrm{Acc}(\tau_i^{k}) 
- \mathrm{Conf}(\tau_i^{k}).
\end{aligned}
\end{equation}
This formulation synthesizes both signals to effectively differentiate between qualitatively distinct reasoning states:
\begin{itemize}
    \item \textbf{Exploration ($\Phi \approx 0$).} Typically observed in the early-to-middle stages of reasoning, where confidence remains low regardless of correctness, reflecting exploratory progress.
    \item \textbf{Correct Confidence ($\Phi \to +1$).} Usually emerging in the later stage once the model has reached the ground-truth answer, characterized by high correctness and high confidence toward the correct solution.
    \item \textbf{False Confidence ($\Phi \to -1$).} Also a late-stage regime where the model becomes highly confident yet incorrect, indicating a confident commitment to an erroneous partial solution.
\end{itemize}

A naive metric based solely on correctness cannot distinguish a model that is still uncertain and exploring from one that has already become confidently committed to a hallucinated conclusion.

Tracking $\Phi(\tau_i^{k})$ over steps $k = 1, \dots, K$ produces a temporal signal that exposes pathological reasoning behaviors such as Over-Checking.

\paragraph{Distinguishing Solving and Checking Tokens.} We decompose reasoning trajectories into solving and checking phases, distinguishing the transition via step-level potential saturation. A step $k$ is classified as checking if the potential has exceeded a high threshold $\varepsilon_{\text{sat}}$ at any earlier step:
\begin{equation}
\text{IsChecking}(k) \iff \exists\, k' < k \;\text{s.t.}\; \Phi(\tau_i^{k'}) > \varepsilon_{\text{sat}}
\end{equation}

Steps with $\text{IsChecking}(k)=\text{false}$ form the solving phase. All tokens inherit the phase label of their respective steps: tokens within checking steps are designated as \emph{Checking Tokens}, while those in solving steps are classified as \emph{Solving Tokens}. While limited verification can be beneficial, \emph{Over-Checking} is quantified as an excessive accumulation of Checking Tokens, which increases inference cost without improving solution quality. Unless otherwise specified, we fix the saturation threshold to $\varepsilon_{\text{sat}} = 0.9$ in all experiments.

\paragraph{Right-to-Wrong Failures due to Over-Checking.} A more severe failure arises when prolonged checking overturns a previously correct solution. We define a Right-to-Wrong (R2W) Failure as:
\begin{equation}
\begin{aligned}
\text{Right-to-Wrong} \iff {} & \max_k \Phi(\tau_i^{k}) > \varepsilon_{\text{sat}} \\
                   & \land\; R(o_i, y^*) = 0
\end{aligned}
\end{equation}
This definition captures cases where the model reaches potential saturation (effectively solving the task) but later overturns the correct solution during prolonged, low-quality self-verification, ending with an incorrect final answer.

To validate Step Potential as a reliable diagnostic signal, we conduct pilot analyses to (i) verify that Step Potential saturation aligns with true solution completion, (ii) test whether truncating at saturation reduces Over-Checking and Right-to-Wrong failures, and (iii) assess the statistical stability of intermediate confidence and correctness. Detailed results are in Appendix~\ref{app:reliability}.

\begin{table*}[t]
\centering
\small
\renewcommand{\arraystretch}{1.10} 
\resizebox{\textwidth}{!}{%
\begin{tabular}{lcccccccccccccccc}
\toprule
 & \multicolumn{2}{c}{\textbf{AIME24}} 
 & \multicolumn{2}{c}{\textbf{AIME25}} 
 & \multicolumn{2}{c}{\textbf{AMC23}} 
 & \multicolumn{2}{c}{\textbf{Minerva}} 
 & \multicolumn{2}{c}{\textbf{Olympiad}} 
 & \multicolumn{2}{c}{\textbf{GPQA}} 
 & \multicolumn{2}{c}{\textbf{Avg.}} \\
\cmidrule(lr){2-3} \cmidrule(lr){4-5} \cmidrule(lr){6-7}
\cmidrule(lr){8-9} \cmidrule(lr){10-11} \cmidrule(lr){12-13}
\cmidrule(lr){14-15}
\textbf{Method} 
 & \textbf{Acc$\uparrow$} & \textbf{Len$\downarrow$}
 & \textbf{Acc$\uparrow$} & \textbf{Len$\downarrow$}
 & \textbf{Acc$\uparrow$} & \textbf{Len$\downarrow$}
 & \textbf{Acc$\uparrow$} & \textbf{Len$\downarrow$}
 & \textbf{Acc$\uparrow$} & \textbf{Len$\downarrow$}
 & \textbf{Acc$\uparrow$} & \textbf{Len$\downarrow$}
 & \textbf{Acc$\uparrow$} & \textbf{Len$\downarrow$} \\
\midrule

\rowcolor{gray!12}
\multicolumn{15}{c}{\textit{Base Model: DeepSeek-R1-Distill-Qwen-7B}} \\
Base            & 52.71 & 13,229 & 39.37 & 14,300 & 90.00 & 6,179 & 57.49 & 4,837 & 72.17 & 8,578 & 54.88 & 7,016 & 61.10 & 9,023 \\
DAST*           & 54.37 & 13,151 & 38.54 & 14,037 & 89.69 & 5,726 & 55.63 & 4,786 & 70.89 & 8,220 & 52.68 & 8,608 & 60.30 & 9,088 \\
LC-R1*          & 49.38 & 7,121 & 34.58 & 8,147 & 87.50 & 2,768 & 54.71 & 1,541 & 67.26 & 4,133 & 52.27 & 3,879 & 57.62 & 4,598 \\
Entropy         & \underline{58.54} & 11,740 & 41.04 & 12,343 & 91.09 & 5,965 & 58.18 & 5,467 & \textbf{74.16} & 7,984 & 55.04 & 7,213 & \underline{63.01} & 8,452 \\
KTAE            & 48.54 & 11,739 & 37.29 & 12,604 & 89.84 & 5,526 & 56.59 & 4,048 & 69.59 & 7,660 & 53.11 & 6,529 & 59.16 & 8,018 \\
DAPO            & 56.25 & 11,245 & \underline{41.46} & 12,722 & \underline{91.72} & 5,772 & \underline{58.30} & 5,107 & 73.12 & 7,744 & \underline{55.52} & 6,691 & 62.73 & 8,213 \\
RF-B           & 56.67 & 11,026 & 40.62 & 12,032 & 90.78 & 5,859 & 58.25 & 4,845 & 73.21 & 7,458 & 54.95 & 6,562 & 62.41 & 7,964 \\
\hdashline
\textbf{SPAE} 
                & \textbf{59.38} & 9,908 & \textbf{42.71} & 10,687 & \textbf{92.50} & 4,543 & \textbf{58.64} & 3,692 & \underline{73.97} & 6,509 & \textbf{55.94} & 5,611 & \textbf{63.86} & 6,825 \\

\midrule
\rowcolor{gray!12}
\multicolumn{15}{c}{\textit{Base Model: DeepSeek-R1-Distill-Llama-8B}} \\
Base            & 44.58 & 13,826 & 28.96 & 14,429 & 87.66 & 7,095 & 43.15 & 5,805 & 65.50 & 9,040 & 53.31 & 7,832 & 53.86 & 9,671 \\
DAPO            & \underline{53.75} & 13,316 & \textbf{37.08} & 14,130 & \underline{92.50} & 7,634 & \underline{50.21} & 7,895 & 71.29 & 7,634 & 55.98 & 8,183 & \underline{60.14} & 9,799 \\
RF-B            & 51.25 & 11,928 & 36.67 & 12,821 & 92.03 & 6,658 & \textbf{50.44} & 6,984 & 71.06 & 8,463 & 55.26 & 7,944 & 59.45 & 9,133 \\
\hdashline
\textbf{SPAE} 
                & \textbf{53.96} & 11,468 & \underline{36.88} & 12,030 & \textbf{92.97} & 5,384 & 49.82 & 5,169 & \textbf{71.39} & 6,996 & \textbf{56.25} & 7,066 & \textbf{60.21} & 8,019 \\

\midrule
\rowcolor{gray!12}
\multicolumn{15}{c}{\textit{Base Model: Qwen3-4B-Thinking}} \\
Base            & \underline{71.04} & 14,375 & 63.96 & 17,266 & \underline{94.84} & 8,086 & 61.35 & 6,575 & \textbf{79.55} & 10,388 & \textbf{51.17} & 8,729 & \underline{70.32} & 10,903 \\
DAPO            & \underline{71.04} & 10,358 & \underline{64.79} & 11,891 & 93.75 & 5,787 & \underline{61.53} & 4,882 & 79.07 & 7,287 & 49.93 & 6,271 & 70.02 & 7,746 \\
RF-B            & 68.96 & 10,489 & 61.67 & 12,278 & 94.06 & 5,874 & \textbf{61.65} & 4,717 & 79.11 & 7,294 & \underline{50.93} & 6,443 & 69.40 & 7,849 \\
\hdashline
\textbf{SPAE} 
                & \textbf{71.88} & 9,608 & \textbf{65.21} & 11,680 & \textbf{96.09} & 5,218 & 61.28 & 4,365 & \underline{79.28} & 6,655 & 50.86 & 5,975 & \textbf{70.77} & 7,250 \\

\bottomrule
\end{tabular}%
}
\caption{Performance comparison of SPAE with various baselines over 16 evaluations. Acc means accuracy(\%) and Len represents the average response length. The best results are in \textbf{bold} and the second-best are \underline{underlined}. ``*'' denotes results obtained by evaluating the official open-source checkpoints.}
\label{tab:main_avg_len}
\vspace{-8pt}
\end{table*}

\begin{figure*}[t]
  \includegraphics[width=\linewidth]{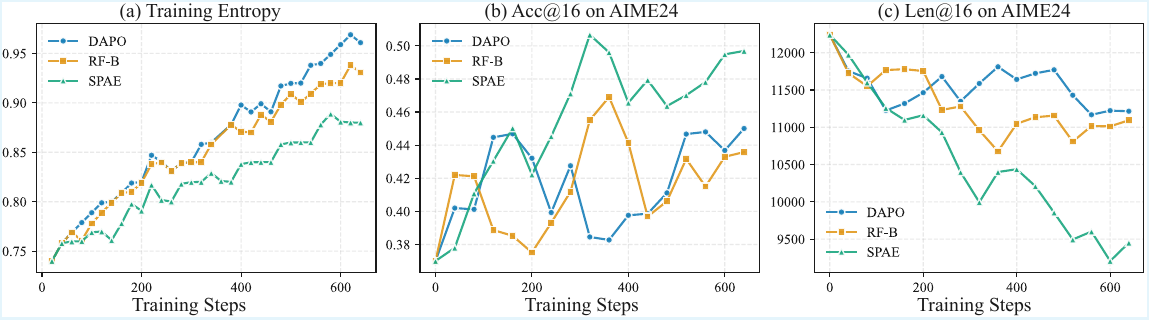}
  \caption {The metric curves of (a) generation entropy during training, (b) test accuracy, and (c) mean response length of DAPO, RF-B and SPAE based on DeepSeek-R1-Distill-Qwen-7B.}
  \label{fig:train_dynamic}
  \vspace{-10pt}
\end{figure*}

\subsection{Step Potential Advantage Estimation}

We propose \emph{Step Potential Advantage Estimation} (SPAE) to incorporate the Step Potential signal from our training-free probe directly into policy optimization. SPAE has two complementary components: \emph{Potential Saturation Penalty}, which downweights the outcome credit after Step Potential saturates to suppress redundant post-solution checking, and \emph{Potential Difference Shaping}, which provides dense step-wise feedback by rewarding progress-inducing transitions and penalizing regressions.

To make the roles of the two components explicit, we write the token-level advantage as
\begin{equation}
\hat{A}_{i,j}^{\text{SPAE}}
=
\underbrace{
\hat{A}_i^{\text{Group}} \cdot f\!\left(\Phi_{i,\mathcal{M}(j)}\right)
}_{\textbf{Saturation Penalty}}
\;+\;
\underbrace{
\xi \cdot g\!\left(\Delta \Phi_{i,\mathcal{M}(j)}\right)
}_{\textbf{Difference Shaping}}
\label{eq:spae_unified}
\end{equation}
where $\xi$ controls the strength of the shaping term.

\paragraph{Step-to-Token Alignment.}
As probing is step-level but optimization is token-level, we define a mapping $\mathcal{M}(j)\in\{1,\dots,K\}$ from token index $j$ to its reasoning step $k$. All tokens in step $\tau_i^{k}$ share the same saturation penalty factor $f(\Phi_{i,k})$ and the same shaping signal $g(\Delta\Phi_{i,k})$.

\paragraph{Potential Saturation Penalty.}
To mitigate Over-Checking, we downweight the outcome advantage once the trajectory has entered the post-saturation regime.
We define the count $C_{\text{sat}}^{(i,k)}$ of preceding saturated steps using an indicator function:
\begin{equation}
C_{\text{sat}}^{(i,k)}
\;=\;
\sum_{t=1}^{k-1}
\mathbb{I}\!\left[\Phi(\tau_i^{t}) > \varepsilon_{\text{sat}} \right],
\label{eq:nsat_indicator}
\end{equation}
Based on this count, we define the saturation penalty factor for step $k$ as
\begin{equation}
f\!\left(\Phi_{i,k}\right)
\;=\;
1 - \alpha \Bigl(1 - \exp\!\bigl(-C_{\text{sat}}^{(i,k)}\bigr)\Bigr),
\label{eq:f_phi}
\end{equation}
which decays initially slowly and then rapidly from $1$ to $1-\alpha$ as saturated steps accumulate.

\paragraph{Potential Difference Shaping.}
We provide step-wise feedback by quantifying the marginal contribution via Step Potential differences:
\begin{equation}
\Delta \Phi_{i,k}
\;=\;
\Phi(\tau_i^{k}) - \Phi(\tau_i^{k-1}),
\end{equation}
Let $\Delta \tilde{\Phi}_{i,k}$ be the Min--Max normalized value of $\Delta \Phi_{i,k}$ within the training batch $\mathcal{B}$.
The shaping function is defined as an exponentially-amplified and batch-centered signal:
\begin{equation}
g\!\left(\Delta \Phi_{i,k}\right)
=
\exp\!\left(\Delta \tilde{\Phi}_{i,k}\right)
-
\mathbb{E}_{\mathcal{B}}
\Bigl[\exp\!\left(\Delta \tilde{\Phi}\right)\Bigr],
\label{eq:shaping_g}
\end{equation}
This term highlights pivotal ``Aha!'' transitions (large positive $\Delta\Phi_{i,k}$) while suppressing trivial steps, and assigns negative contributions to relative regressions after batch-centering.

\paragraph{Integrated Advantage Estimation.}
Finally, we compute the group-relative outcome advantage without standard deviation:
\begin{equation}
\hat{A}_i^{\text{Group}} = R_i - \operatorname{mean}(\{R_k\}_{k=1}^G),
\end{equation}
After computing $\hat{A}_{i,j}^{\text{SPAE}}$ in Eq.~\ref{eq:spae_unified}, we apply global batch advantage normalization over all tokens in the training batch $\mathcal{B}$ for stability:
\begin{equation}
\hat{A}_{i,j}^{\text{Final}} =
\frac{
\hat{A}_{i,j}^{\text{SPAE}} - \operatorname{mean}\!\left(\left\{\hat{A}^{\text{SPAE}} \mid \hat{A}^{\text{SPAE}}\in\mathcal{B}\right\}\right)
}{
\operatorname{std}\!\left(\left\{\hat{A}^{\text{SPAE}} \mid \hat{A}^{\text{SPAE}}\in\mathcal{B}\right\}\right) + \epsilon
}.
\end{equation}
The overall SPAE algorithm is summarized in Algorithm~\ref{alg:spae}.

\begin{table*}[t]
\centering
\small
\renewcommand{\arraystretch}{1} 
\setlength{\tabcolsep}{4.5pt}      
\resizebox{\textwidth}{!}{%
\begin{tabular}{l cccccc cc}
\toprule
\multirow{2}{*}{\textbf{Setting}} & \textbf{AIME24} & \textbf{AIME25} & \textbf{AMC23} & \textbf{Minerva} & \textbf{Olympiad} & \textbf{GPQA} & \textbf{Avg.} & \textbf{Avg.} \\
& \textbf{Acc} & \textbf{Acc} & \textbf{Acc} & \textbf{Acc} & \textbf{Acc} & \textbf{Acc} & \textbf{Acc} & \textbf{Len} \\
\midrule

\textbf{SPAE (Full)}
& \underline{59.38} & \underline{42.71} & \textbf{92.50} & \underline{58.64} & \textbf{73.97} & \textbf{55.94} & \underline{63.86} & 6,825 \\

\quad w/o Conf in Potential
& 58.54 & 40.62 & 91.56 & 58.27 & 71.39 & 54.20 & 62.43 & 6,967 \\

\quad w/o Difference Shaping
& 57.71 & 39.79 & 91.87 & 57.95 & 72.71 & 54.80 & 62.47 & \underline{6,773} \\

\quad w/o Saturation Penalty
& \textbf{61.46} & \textbf{42.92} & 91.72 & \textbf{58.92} & \underline{73.88} & 54.98 & \textbf{63.98} & 7,517 \\

Baseline (RF-B)
& 56.67 & 40.62 & 90.78 & 58.25 & 73.21 & 54.95 & 62.41 & 7,964 \\

\hdashline \noalign{\vskip 2pt}

Shaping Factor ($\xi=0.1$)
& 53.54 & 42.50 & 91.25 & 58.59 & 73.19 & 55.34 & 62.40 & 6,866 \\

\phantom{Shaping Factor} ($\xi=1.0$)
& 57.29 & 42.50 & 91.41 & 58.41 & 72.71 & 55.32 & 62.94 & 6,768 \\

Penalty Factor ($\alpha=0.1$)
& 56.67 & \underline{42.71} & \underline{92.66} & 58.23 & 73.79 & \underline{55.54} & 63.27 & 7,348 \\

\phantom{Penalty Factor} ($\alpha=1.0$)
& 57.50 & 40.42 & 91.87 & 58.18 & 73.01 & 55.11 & 62.68 & \textbf{6,183} \\

\bottomrule
\end{tabular}%
}
\caption{\textbf{Ablation and Sensitivity Analysis on DeepSeek-R1-Distill-Qwen-7B.} We report accuracy (\%) on individual benchmarks, plus the average accuracy and response length across all tasks over 16 evaluations. ``SPAE (Full)'' uses $\alpha=0.5, \xi=0.5$. The best results are in \textbf{bold} and the second-best are \underline{underlined}.}
\label{tab:ablation}
\vspace{-6pt}
\end{table*}

\begin{figure}[t]
    \centering
    \includegraphics[width=\linewidth]{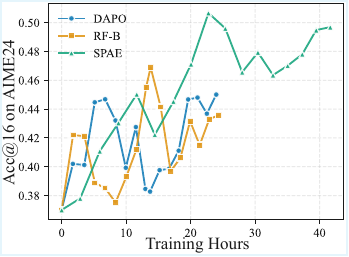}
    \caption{Training efficiency on AIME2024: accuracy vs.\ cumulative training hours. SPAE reaches higher accuracy under the same wall-clock budget.}
    \label{fig:training_efficiency}
    \vspace{-6pt}
\end{figure}

\section{Experiments}

\subsection{Setup}

\paragraph{Models and Baselines.} We train DeepSeek R1-Distill-Qwen-7B, R1-Distill-Llama-8B \citep{guo2025deepseek}, and Qwen3-4B \citep{yang2025qwen3technicalreport} on DAPO-MATH-17K \citep{yu2025dapo}. For all backbones, we train and report results for \textbf{DAPO} \citep{yu2025dapo} and \textbf{RF-B} \citep{hu2025reinforce} as our RLVR baselines. For the 7B model, we further compare two token-level extensions built upon the DAPO training pipeline: \textbf{KTAE} \citep{sun2025ktae}, which assigns token-level advantages by combining rollout outcomes with statistical token-importance estimates, and \textbf{Entropy Advantage} \citep{cheng2025reasoningexplorationentropyperspective}, which augments advantages with an entropy-based intrinsic term. We evaluate efficient reasoning baselines \textbf{DAST} \citep{shen-etal-2025-dast} and \textbf{LC-R1} \citep{cheng2025optimizing} by directly employing official open-sourced checkpoints without retraining. All our training runs use VeRL \citep{sheng2025hybridflow} with an off-policy setup (global batch 640, mini-batch 32). For SPAE, we set $\xi=\alpha=0.5$ and $N=5$.

\paragraph{Evaluation.} We evaluate on in-domain math benchmarks (AIME2024 \& 2025, AMC2023, Minerva-Math \citep{Lewkowycz2022}, OlympiadBench \citep{he-etal-2024-olympiadbench}) and the out-of-domain GPQA \citep{rein2024gpqa}. For answer verification, we use the \textsc{Math-Verify} library together with xVerify-3B-Ia verifier model \citep{chen2025xverifyefficientanswerverifier} to robustly check final answers. Decoding uses temperature 0.6, top-$k$ 50, top-$p$ 1.0, and a max length of 32{,}768 tokens. We report \textbf{Acc@16} (mean accuracy over 16 runs) and \textbf{Len@16} (mean generated tokens over 16 runs). Additional details are provided in Appendix~\ref{app:experiment_details}.

\subsection{Main Results}

Table~\ref{tab:main_avg_len} compares SPAE with RLVR baselines (DAPO, RF-B), efficiency methods (DAST, LC-R1), and token-level shaping (KTAE, Entropy). Across three backbones, SPAE yields the best accuracy--length trade-off: SPAE improves accuracy while consistently shortening responses.

\paragraph{Accuracy.}
SPAE achieves the best average accuracy across all backbones. On Qwen-7B, SPAE reaches \textbf{63.86\%}, surpassing DAPO (62.73\%) and RF-B (62.41\%). On Llama-8B, SPAE boosts the base model by \textbf{+6.35\%} (53.86\% $\rightarrow$ 60.21\%), outperforming all other baselines. Notably, on Qwen3-4B, SPAE is the only method that improves over the base model (+0.45\%), while other RLVR baselines regress. In contrast, efficient reasoning methods (DAST and LC-R1) consistently suffer from accuracy drops relative to the base model. Furthermore, the advantage estimation baseline KTAE fails to scale effectively with long CoT reasoning, resulting in overall performance degradation.

\paragraph{Efficiency.}
SPAE substantially reduces generation length by diminishing the advantage of post-solution segments in correct responses. Relative to the base models, SPAE achieves average token usage reductions of \textbf{$\sim$24\%} on Qwen-7B, \textbf{$\sim$17.1\%} on Llama-8B, and \textbf{$\sim$33.5\%} on Qwen3-4B. In contrast to strict length-constrained training (LC-R1), which shortens outputs at the cost of notable accuracy drops, SPAE improves correctness while compressing length. Furthermore, on the OOD GPQA dataset, SPAE maintains competitive performance despite the reduced inference cost, demonstrating robust generalization capabilities.

\paragraph{Training Dynamics.}
Figure~\ref{fig:train_dynamic} shows training curves on AIME2024 for R1-Distill-Qwen-7B (other backbones in Appendix~\ref{sec:ext_results}). SPAE steadily reduces response length throughout training, whereas DAPO and RF-B plateau at much longer generations. This compression co-occurs with higher test accuracy and lower entropy growth, suggesting that step-potential shaping provides a more stable credit signal than purely group-relative baselines. Despite the extra probing cost, SPAE achieves better accuracy under the same wall-clock budget (Figure~\ref{fig:training_efficiency}).

\subsection{Ablation Study}

\paragraph{Component Necessity.}
Table~\ref{tab:ablation} validates the necessity of each component across benchmarks on R1-Distill-Qwen-7B. We first test an Acc-only potential variant by discarding confidence (w/o Conf in Potential). This variant underperforms full SPAE (62.43\% vs.\ 63.86\%, $\mathbf{-1.43}$) and produces longer responses (6,967 vs.\ 6,825, $\mathbf{+142}$). This shows that correctness alone cannot reliably distinguish uncertain exploration from confident errors, leading to noisier step transitions. Removing Potential Difference Shaping ($\xi=0$) reduces the average accuracy from 63.86\% to 62.47\% ($\mathbf{-1.39}$), suggesting its key role in guiding step-wise progress. In contrast, removing the Potential Saturation Penalty ($\alpha=0$) yields a negligible change in average accuracy (63.98\%, +0.12) but increases the average response length from 6,825 to 7,517 tokens ($\mathbf{+692}$), indicating the penalty term $f(\Phi)$ primarily curbs Over-Checking and improves efficiency.

\paragraph{Sensitivity.}
Sweeping $\xi,\alpha \in \{0.1,0.5,1.0\}$ shows SPAE is robust and achieves its best accuracy--efficiency trade-off at $\xi=\alpha=0.5$. Weak shaping ($\xi=0.1$) provides insufficient guidance, reducing Avg.\ Acc to 62.40\% ($-1.46$), while overly strong shaping ($\xi=1.0$) also hurts (62.94\%, $-0.92$). For redundancy control, a mild penalty ($\alpha=0.1$) does not sufficiently prune post-saturation computation (Avg.\ Len 7,348; $\mathbf{+523}$), whereas an aggressive penalty ($\alpha=1.0$) over-truncates (Avg.\ Len 6,183; $-642$) and degrades accuracy to 62.68\% ($-1.18$).

\subsection{Analysis of Reasoning Behaviors}
Table~\ref{tab:diagnostic_dynamics_compact} summarizes Over-Checking statistics on correct trajectories and the R2W failure rate on incorrect trajectories of R1-Distill-Qwen-7B. 

Baselines exhibit substantial Over-Checking, spending 1.3K--1.8K checking tokens after saturation (e.g., 1,511 in Base and 1,787 in DAPO). SPAE mitigates this by surgically reducing checking to \textbf{614} tokens ($-59\%$ vs.\ Base) while retaining a robust solving budget (3,483), achieving the best accuracy (\textbf{51.05\%}). Moreover, SPAE reduces reflective behaviors on correct trajectories (Reflect: 9.08 vs.\ 17.72 in Base), suggesting it discourages unnecessary post-solution self-reflection without sacrificing correctness. In contrast, LC-R1 shortens both Solve and Check aggressively (2,828 / 299), which correlates with a large accuracy drop, indicating indiscriminate truncation.

Baselines suffer high R2W rates (e.g., 8.10\% for Base and 10.31\% for DAST). SPAE substantially reduces R2W to \textbf{2.65\%} ($-67\%$ vs.\ Base), indicating that suppressing redundant checking also prevents destructive self-correction that flips a previously correct intermediate conclusion.

\begin{table}[t]
\centering
\small
\renewcommand{\arraystretch}{1.05}
\setlength{\tabcolsep}{4.2pt}
\resizebox{\columnwidth}{!}{%
\begin{tabular}{lccccc}
\toprule
\textbf{Method} & \textbf{Acc} & \textbf{Solve} & \textbf{Check} & \textbf{Reflect} & \textbf{R2W} \\
\midrule
Base          & 46.04 & 4,354 & 1,511 & 17.72 & 8.10 \\
DAST          & 46.46 & 3,970 & 1,424 & 16.56 & 10.31 \\
LC-R1         & 41.98 & 2,828 &   299 &  5.53 & 4.67 \\
DAPO          & 48.86 & 4,581 & 1,787 & 19.05 & 9.50 \\
RF-B          & 48.65 & 4,414 & 1,313 & 15.58 & 6.29 \\
\hdashline
\textbf{SPAE} & \textbf{51.05} & 3,483 & 614 &  9.08 & \textbf{2.65} \\
\bottomrule
\end{tabular}%
}
\caption{\textbf{Reasoning behaviors on AIME2024 \& 2025.} Solve/Check/Reflect are measured on \emph{correct} trajectories: Solve/Check denote token lengths in the solving and checking phases, and Reflect counts steps containing explicit self-reflective tokens (e.g., ``wait'', ``alternatively''). R2W is the Right-to-Wrong failure rate on \emph{incorrect} trajectories.}

\label{tab:diagnostic_dynamics_compact}
\vspace{-6pt}
\end{table}

\section{Conclusion}

In this paper, we mitigate the credit assignment ambiguity in RLVR by introducing SPAE. By leveraging a training-free probing mechanism, we formalize Step Potential to explicitly quantify reasoning progress, allowing us to identify and mitigate pathological behaviors such as Over-Checking and Right-to-Wrong failures. Unlike previous token-level or length-penalty approaches, SPAE provides dense, step-aware supervision that aligns policy optimization with semantic convergence. Our extensive experiments across Qwen and Llama families demonstrate that SPAE achieves a superior Pareto frontier between performance and cost: it significantly boosts accuracy on challenging benchmarks while reducing inference latency through the precise pruning of redundant verification steps.

\section*{Limitations}

SPAE introduces additional computation during training. Our future work will explore more lightweight or adaptive probing strategies.

The correctness probe currently assumes structured answers, which limits applicability to free-form outputs. Developing format-agnostic correctness estimators is an important direction.

Our experiments focus on mathematical reasoning. Extending SPAE to code generation and broader reasoning domains remains future work.


\bibliography{custom}

\appendix

\section{LLM Usage}

We used LLMs only to polish grammar and improve the clarity of the manuscript. All research ideas, experiments, and analyses were conducted by the authors.

\section{Related Work}

\paragraph{RL for LLM Reasoning.}
Reinforcement learning has shifted from preference alignment toward becoming a primary driver of complex reasoning in LLMs. Early reasoning-oriented alignment largely relied on Proximal Policy Optimization (PPO) \citep{schulman2017ppo}, but the need to train and maintain a separate value network makes PPO increasingly costly for long CoT reasoning. This has accelerated adoption of RLVR, where policy updates are guided by outcome-level correctness without a learnable critic. Representative methods include ReMax \citep{li2024remax}, Reinforce++-Baseline \citep{hu2025reinforce}, and GRPO \citep{shao2024deepseekmath}, which estimate advantages using group- or batch-based baselines and enable scalable long-CoT training. Despite their practicality, these approaches still rely on sparse terminal rewards, providing limited guidance for identifying which intermediate reasoning steps are truly helpful and which are redundant or harmful.

\paragraph{Token-Level Advantage Estimation.}
Recent work has explored fine-grained credit assignment to improve RLVR. DAPO \citep{yu2025dapo} improves training stability via token-averaged objectives and global normalization, but it still lacks explicit supervision of individual tokens or steps. Consequently, entropy-based estimation have emerged as a common proxy for token importance: entropy-guided shaping scales gradients based on token uncertainty under the premise that high-entropy tokens correspond to critical branching points \citep{cheng2025reasoningexplorationentropyperspective, chen2025seedgrposemanticentropyenhanced, le2025promptleftbehindexploiting, wang2025harnessinguncertaintyentropymodulatedpolicy}. These methods often upweight high-entropy tokens in correct trajectories to encourage reflective computation, while tempering penalties on high-entropy tokens in incorrect trajectories to preserve exploration. Related ideas also modulate gradients using planning-related tokens \citep{wang2025emergenthierarchicalreasoningllms} or statistical correlates of token importance (e.g., \citealt{sun2025ktae}). Our work targets a missing ingredient: a semantically grounded, step-level estimate of reasoning progress that can distinguish essential deduction from redundancy.

\paragraph{Efficient Reasoning Techniques.}
A complementary line of work focuses on reducing inference-time compute. Length-aware objectives fine-tune models with length preferences to encourage shorter reasoning traces while maintaining correctness, enabling ``long-to-short'' behavior and improving efficiency \citep{shen-etal-2025-dast, cheng2025optimizing, aggarwal2025l}. Other approaches train models to adaptively decide whether to produce a chain-of-thought based on problem difficulty, typically via a two-stage pipeline combining supervised fine-tuning and RL so that the model learns when extended reasoning is necessary \citep{zhang-etal-2025-adaptthink, lou2025adacotparetooptimaladaptivechainofthought, xiong2025mixturereasoningsteachlarge}. Similar ideas appear in early Qwen3 \citep{yang2025qwen3technicalreport} deployments but may require user-side control. While effective for reducing average length, these methods still largely treat the model output as a single sequence-level object during training and do not explicitly separate efficient reasoning segments from inefficient ones. Our approach addresses this gap by making step-level progress observable and using it to shape credit assignment and suppress redundant post-solution checking directly within RL optimization.

\section{Reliability of Step Potential as a Diagnostic Signal}
\label{app:reliability}

This section examines whether Step Potential can serve as a reliable diagnostic signal for reasoning dynamics. Unless otherwise specified, all analyses are conducted with DeepSeek-R1-Distill-Qwen-7B on 60 problems from AIME2024 \& 2025, using 16 sampled responses per problem (i.e., @16 evaluation). We validate Step Potential from three angles: (i) temporal alignment with oracle supervision, (ii) the impact of forced truncation on correctness (validating Over-Checking), and (iii) the statistical stability of the estimator.

\begin{table}[t]
\centering
\small
\setlength{\tabcolsep}{6pt}
\renewcommand{\arraystretch}{1.05}
\begin{tabular}{lc}
\toprule
\textbf{Statistic} & \textbf{Value} \\
\midrule
$\Pr(\Delta k = 0)$ & 86.0\% \\
$\Pr(\Delta k > 0)$ & 3.5\% \\
$\Pr(\Delta k < 0)$ & 10.5\% \\
$\mathbb{E}[|\Delta k|]$ & 3.86 \\
\bottomrule
\end{tabular}
\caption{Temporal alignment between probe saturation and oracle supervision.}
\label{tab:alignment_results}
\end{table}

\subsection{Temporal Alignment with Oracle Supervision}
\label{app:alignment}

We first test whether the first step where Step Potential saturates corresponds to the earliest moment when the model has already formed a complete correct solution. To obtain an oracle reference, we employ Qwen3-235B-A22B-Instruct-2507 \citep{yang2025qwen3technicalreport} to retrospectively inspect each trajectory with the prompt in Figure~\ref{fig:annotation_prompt}. The teacher model identifies the exact text span where the correct logic and answer are first fully established. We map this boundary back to our step index, denoting it as the Ground Truth Solving Step $k_{\text{GT}}$.

\paragraph{Metric.}
We define the probe-detected solving step $k_{\text{Probe}}$ as the earliest step exceeding the saturation threshold $\varepsilon_{\text{sat}}$:
\begin{equation}
    k_{\text{Probe}} = \min \{k : \Phi(\tau_i^{k}) > \varepsilon_{\text{sat}} \}.
\end{equation}
We then measure the step displacement $\Delta k = k_{\text{Probe}} - k_{\text{GT}}$, where $\Delta k=0$ implies perfect synchronization; $\Delta k > 0$ implies delayed detection; and $\Delta k < 0$ implies early triggering.

\paragraph{Results.}
As shown in Table~\ref{tab:alignment_results}, Step Potential saturation exhibits strong temporal agreement with oracle supervision: \textbf{86.0\%} of trajectories achieve exact synchronization ($\Delta k=0$), indicating that the probe typically triggers at the same step boundary where the oracle judges the full correct solution to be established. For the remaining cases with $\Delta k\neq 0$, early triggering dominates: \textbf{75.0\%} of non-zero displacements satisfy $\Delta k<0$, whereas only \textbf{25.0\%} are delayed detections ($\Delta k>0$). The mean absolute displacement is $\mathbb{E}[|\Delta k|]=3.86$, suggesting that mismatches, while infrequent, can span a few step boundaries; qualitatively, these cases often correspond to partially implicit derivations where the model has already converged to the correct answer (high confidence and correctness) before the teacher deems the full reasoning to be explicitly complete. Overall, these results support Step Potential saturation as a practical marker for separating solving from post-solution checking.

\subsection{The Oracle Truncation Test: Validating Over-Checking}
\label{app:truncation}

To confirm that steps generated after saturation are largely redundant and potentially harmful, we perform an intervention experiment.

\paragraph{Protocol.}
We compare standard decoding against a Probe-Truncated Decoding strategy:
\begin{enumerate}
    \item \textbf{Monitor:} At each step boundary $k$, we compute $\Phi(\tau_i^{k})$ using the probing mechanism.
    \item \textbf{Intervene:} If $\Phi(\tau_i^{k}) > \varepsilon_{\text{sat}}$, we immediately append the \texttt{</think>} token to close the reasoning block.
    \item \textbf{Output:} The model is then forced to generate the final summary $s$, discarding any subsequent reasoning steps that would have been generated.
\end{enumerate}

\paragraph{Results.}
As summarized in Table~\ref{tab:truncation_results}, oracle truncation yields clear efficiency gains while improving reliability: probe-truncated decoding reduces the average output length (Len@16) from \textbf{13,765} to \textbf{12,931}, and simultaneously increases Acc@16 by \textbf{2.40 points} from \textbf{46.04} to \textbf{48.44}. Notably, truncation eliminates R2W failures entirely, driving the R2W rate down from \textbf{5.4} to \textbf{0.0}. These results directly support the Over-Checking hypothesis: once Step Potential saturates, continued generation is largely redundant and can even induce spurious self-contradictions that overwrite an already-correct solution.

\begin{table}[t]
\centering
\small
\setlength{\tabcolsep}{5pt}
\renewcommand{\arraystretch}{1.05}
\begin{tabular}{lccc}
\toprule
\textbf{Method} & \textbf{Len@16} & \textbf{Acc@16} & \textbf{R2W} \\
\midrule
Standard Decoding & 13765 & 46.04 & 5.4 \\
Probe-Truncated   & 12931 & 48.44 & \textbf{0.0} \\
\bottomrule
\end{tabular}
\caption{Comparison between standard decoding and probe-truncated decoding. Truncation effectively eliminates Right-to-Wrong (R2W) failures caused by Over-Checking.}
\label{tab:truncation_results}
\end{table}

\begin{table}[t]
\centering
\small
\setlength{\tabcolsep}{8pt}
\renewcommand{\arraystretch}{1.05}
\begin{tabular}{lcc}
\toprule
\textbf{Step Progress Bin} & $\overline{\mathrm{Var}}[\mathrm{Conf}]$ & $\overline{\mathrm{Var}}[\mathrm{Acc}]$ \\
\midrule
$[0,0.2)$   & 0.00216 & 0.00341 \\
$[0.2,0.4)$ & 0.00341 & 0.00395 \\
$[0.4,0.6)$ & 0.00402 & 0.00377 \\
$[0.6,0.8)$ & 0.00393 & 0.00318 \\
$[0.8,1.0]$ & 0.00157 & 0.00279 \\
\bottomrule
\end{tabular}
\caption{Progress-conditioned sampling variance of probe signals. For each step, we compute the within-step variance across 16 probe samples for $\mathrm{Conf}$ and $\mathrm{Acc}$, then report the mean variance in each relative-progress bin.}
\label{tab:variance_bins}
\end{table}

\subsection{Variance and Stability Analysis}
\label{app:variance}

Since Step Potential is derived from stochastic probe sampling, we analyze the stability of its underlying components---\textbf{confidence} and \textbf{correctness}---by measuring their \emph{within-step} sampling variance across probe continuations.

\paragraph{Protocol.}
For each response, we segment the reasoning trajectory into $K$ steps and run the probing mechanism at every step boundary to obtain $N=16$ probe continuations. For each step $k$, the probe yields per-sample estimates $\{\mathrm{Conf}^{(n)}_k\}_{n=1}^{16}$ and $\{\mathrm{Acc}^{(n)}_k\}_{n=1}^{16}$, from which we compute the within-step sampling variance:
\begin{equation}
\begin{aligned}
\mathrm{Var}_{\text{probe}}\!\left[\mathrm{Conf}_k\right]
&= \mathrm{Var}\!\left(\{\mathrm{Conf}^{(n)}_k\}_{n=1}^{16}\right), \\
\mathrm{Var}_{\text{probe}}\!\left[\mathrm{Acc}_k\right]
&= \mathrm{Var}\!\left(\{\mathrm{Acc}^{(n)}_k\}_{n=1}^{16}\right).
\end{aligned}
\end{equation}

\paragraph{Step-progress Binning.}
To characterize how sampling stability evolves over the course of reasoning, we bin steps by their relative progress $r_k = k/K$ into five intervals:
\[
[0,0.2),\ [0.2,0.4),\ [0.4,0.6),\ [0.6,0.8),\ [0.8,1.0].
\]
For each bin $b$, we aggregate the variances over all steps whose $r_k$ falls into $b$, reporting the mean variance:
\begin{equation}
\begin{aligned}
\overline{\mathrm{Var}}_b[\mathrm{Conf}] &= \mathbb{E}_{k \in b}\!\left[\mathrm{Var}_{\text{probe}}[\mathrm{Conf}_k]\right], \\
\overline{\mathrm{Var}}_b[\mathrm{Acc}]  &= \mathbb{E}_{k \in b}\!\left[\mathrm{Var}_{\text{probe}}[\mathrm{Acc}_k]\right].
\end{aligned}
\end{equation}

\paragraph{Results.}
Table~\ref{tab:variance_bins} shows that the probe signals are statistically stable and exhibit a meaningful dependence on reasoning progress: both $\overline{\mathrm{Var}}[\mathrm{Conf}]$ and $\overline{\mathrm{Var}}[\mathrm{Acc}]$ are smallest near the beginning and end of trajectories, while peaking in the middle bins, consistent with an exploratory phase where the model has not yet converged. In particular, $\overline{\mathrm{Var}}[\mathrm{Conf}]$ increases from \textbf{0.00216} in $[0,0.2)$ to a maximum of \textbf{0.00402} in $[0.4,0.6)$, then drops to \textbf{0.00157} in the final bin $[0.8,1.0]$; $\overline{\mathrm{Var}}[\mathrm{Acc}]$ shows a similar pattern, peaking at \textbf{0.00395} in $[0.2,0.4)$ and decreasing to \textbf{0.00279} in $[0.8,1.0]$. This progress-conditioned variance indicates that the stochasticity of the probe is not arbitrary noise; rather, it peaks when the model exhibits high uncertainty and diminishes as the trajectory stabilizes, thereby establishing a robust foundation for employing Step Potential as both a diagnostic marker and a dense shaping signal.

\section{Experiment Details}
\label{app:experiment_details}

\subsection{Datasets}
We evaluate both in-domain mathematical reasoning and out-of-domain generalization. We use the official test sets and standard answer formats, strictly adhering to the licenses associated with each dataset.

\paragraph{Evaluation Benchmarks.}
\begin{itemize}
    \item \textbf{AIME 2024 (\#30)\footnote{\url{https://huggingface.co/datasets/hendrydong/aime24}} / AIME 2025 (\#30)\footnote{\url{https://huggingface.co/datasets/math-ai/aime25}}.}
    American Invitational Mathematics Examination problems. Answers are typically integers with a fixed format, which supports reliable verification.
    
    \item \textbf{AMC 2023 (\#40)\footnote{\url{https://huggingface.co/datasets/zwhe99/amc23}}.}
    American Mathematics Competitions problems.
    
    \item \textbf{Minerva-Math (\#272).} \citep{Lewkowycz2022}
    A collection of mathematical problems curated for evaluating step-by-step reasoning, covering a wide range of topics and difficulty levels.
    
    \item \textbf{OlympiadBench (text-only EN math subset, \#674).} \citep{he-etal-2024-olympiadbench}
    Olympiad-style problems that emphasize long-horizon symbolic reasoning and composition of multiple lemmas.
    
    \item \textbf{GPQA (\#448).} \citep{rein2024gpqa}
    A challenging question-answering benchmark intended to test out-of-domain generalization and scientific reasoning.
\end{itemize}

\paragraph{Training Data.}
All models are fine-tuned on \textbf{DAPO-MATH-17K} \citep{yu2025dapo}, which consists of 17K prompts, each paired with an integer as the answer.

\subsection{Baselines}

We compare SPAE against strong RLVR baselines, token-level advantage estimation methods, and efficient reasoning approaches. For fair comparison, we match training data, rollout settings, and decoding configurations whenever applicable. When official checkpoints are used, we report results under the authors’ recommended inference settings and additionally evaluate under our standardized decoding protocol when possible.

\subsubsection{RLVR Baselines}
\begin{itemize}
    \item \textbf{DAPO} \citep{yu2025dapo}. A stabilized RLVR variant featuring decoupled clipping bounds, dynamic sampling to maintain reward variance within groups, and global token-level normalization to balance updates across variable-length rollouts.
    \item \textbf{Reinforce++-Baseline} \citep{hu2025reinforce}. An RLVR method that improves stability via global batch advantage normalization, normalizing advantages using statistics over the full training batch to reduce sensitivity to small group sizes and outliers.
\end{itemize}

\subsubsection{Token-Level Advantage Estimation}
\begin{itemize}
    \item \textbf{Entropy Advantage} \citep{cheng2025reasoningexplorationentropyperspective}. This method augments the advantage function with an entropy-based term to encourage exploration.
    \item \textbf{Key Token Advantage Estimation (KTAE)} \citep{sun2025ktae}. KTAE addresses the coarse-grained credit assignment issue in group-based RLVR by estimating token-level importance without additional learned models; it combines rollout-level outcome information with a statistical token-importance signal to enable finer-grained advantage assignment.
\end{itemize}

\subsubsection{Efficient Reasoning Methods}
\begin{itemize}
    \item \textbf{LC-R1} \citep{cheng2025optimizing}. An RL approach for efficient reasoning that incorporates length-aware reward components (e.g., length and compression rewards) in addition to correctness to encourage output compression with minimal accuracy loss.
    \item \textbf{DAST (Difficulty-Adaptive Slow Thinking)} \citep{shen-etal-2025-dast}. A framework that adapts Chain-of-Thought length to problem difficulty via budget-aware reward shaping and preference optimization, penalizing overly long responses on easier instances while preserving sufficient reasoning for hard ones.
\end{itemize}

\begin{table}[t]
\centering
\small
\renewcommand{\arraystretch}{1.10} 
\setlength{\tabcolsep}{4pt}
\resizebox{\columnwidth}{!}{%
\begin{tabular}{lcccccc}
\toprule
\textbf{Method} 
 & \textbf{AIME24} 
 & \textbf{AIME25} 
 & \textbf{AMC23} 
 & \textbf{Minerva} 
 & \textbf{Olympiad} 
 & \textbf{GPQA} \\
\midrule

\rowcolor{gray!12}
\multicolumn{7}{c}{\textit{Base Model: DeepSeek-R1-Distill-Qwen-7B}} \\
Base            & 76.67 & 66.67 & \textbf{100.00} & 79.41 & 87.98 & 91.74 \\
DAST & 83.33 & 70.00 & \textbf{100.00} & 78.68 & 87.83 & 90.40 \\
LC-R1 & 80.00 & 70.00 & 97.50 & 79.41 & 87.39 & \textbf{91.96} \\
Entropy         & 80.00 & 63.33 & 97.50 & 78.31 & 87.83 & 91.07 \\
KTAE            & 80.00 & 70.00 & 97.50 & 79.04 & 86.80 & 91.52 \\
DAPO            & 80.00 & 70.00 & 97.50 & 79.41 & 87.69 & 89.73 \\
RF-B            & 83.33 & \textbf{73.33} & \textbf{100.00} & 78.68 & 87.98 & 89.51 \\
\hdashline
\textbf{SPAE} 
                & \textbf{86.67} & 70.00 & \textbf{100.00} & \textbf{80.15} & \textbf{88.13} & 89.73 \\

\midrule
\rowcolor{gray!12}
\multicolumn{7}{c}{\textit{Base Model: DeepSeek-R1-Distill-Llama-8B}} \\
Base            & 80.00 & 66.67 & 97.50 & 75.37 & 86.50 & \textbf{90.18} \\
DAPO            & \textbf{83.33} & \textbf{70.00} & 97.50 & 78.68 & \textbf{87.39} & 86.83 \\
RF-B            & \textbf{83.33} & 60.00 & \textbf{100.00} & \textbf{79.04} & 86.80 & 87.50 \\
\hdashline
\textbf{SPAE} 
                & \textbf{83.33} & 63.33 & \textbf{100.00} & 78.68 & 86.50 & 87.72 \\

\midrule
\rowcolor{gray!12}
\multicolumn{7}{c}{\textit{Base Model: Qwen3-4B-Thinking}} \\
Base            & \textbf{86.67} & \textbf{83.33} & \textbf{100.00} & 76.10 & \textbf{90.95} & \textbf{78.12} \\
DAPO            & 80.00 & \textbf{83.33} & \textbf{100.00} & 75.37 & 89.02 & 77.01 \\
RF-B            & 83.33 & \textbf{83.33} & \textbf{100.00} & \textbf{76.47} & 89.32 & 77.01 \\
\hdashline
\textbf{SPAE} 
                & 83.33 & \textbf{83.33} & \textbf{100.00} & 75.00 & 89.02 & 76.56 \\

\bottomrule
\end{tabular}%
}
\caption{Pass@16 performance comparison of SPAE with various baselines. Best results in each block are highlighted in \textbf{bold}.}
\label{tab:pass16_results}
\end{table}

\subsection{Training Details}

We use a group-based RLVR training setup across all models.

\begin{itemize}
    \item \textbf{Hardware.} All experiments are conducted on $32 \times$ NVIDIA H200 GPUs.
    \item \textbf{Batching and framework.} Training is implemented in VeRL \citep{sheng2025hybridflow} with an off-policy scheme. We use a global batch size of 640, processed in a mini-batch size of 32.
    
    \item \textbf{Rollout Configuration.} During rollout, the group size is set to $G=8$. The group sampling temperature is 1.0. The maximum sampled length is 16,384 tokens. The system prompt is shown in Figure~\ref{fig:systemprompt}.
    
    \item \textbf{Optimization.} We use a learning rate of $1 \times 10^{-6}$. The KL-divergence regularization term is omitted. We use decoupled clipping bounds with $\varepsilon_{\mathrm{high}} = 0.28$ and $\varepsilon_{\mathrm{low}} = 0.2$, where the higher upper bound encourages diversity and exploration during rollout updates.
    
    \item \textbf{SPAE hyperparameters.} We fix $\xi=0.5, \alpha=0.5$ and $N=5$.
    
    \item \textbf{Training steps.} We train all R1-Distill-Qwen-7B variants for 640 steps, all R1-Distill-Llama-8B variants for 600 steps, and all Qwen3-4B variants for 560 steps.
\end{itemize}

\begin{figure}[t]
\centering
\begin{tcolorbox}[
    colback=white,
    colframe=gray!70,
    arc=4pt,
    boxrule=0.8pt,
    width=0.9\linewidth,
    fontupper=\ttfamily\footnotesize,
    title=\textbf{System Prompt},
    coltitle=white,
    colbacktitle=gray!70
]
Please reason step by step, and put your final answer within \textbackslash boxed\{\}.
\end{tcolorbox}
\caption{The system prompt for training and test.}
\label{fig:systemprompt}
\end{figure}

\begin{figure*}[t]
\centering
\includegraphics[width=\linewidth]{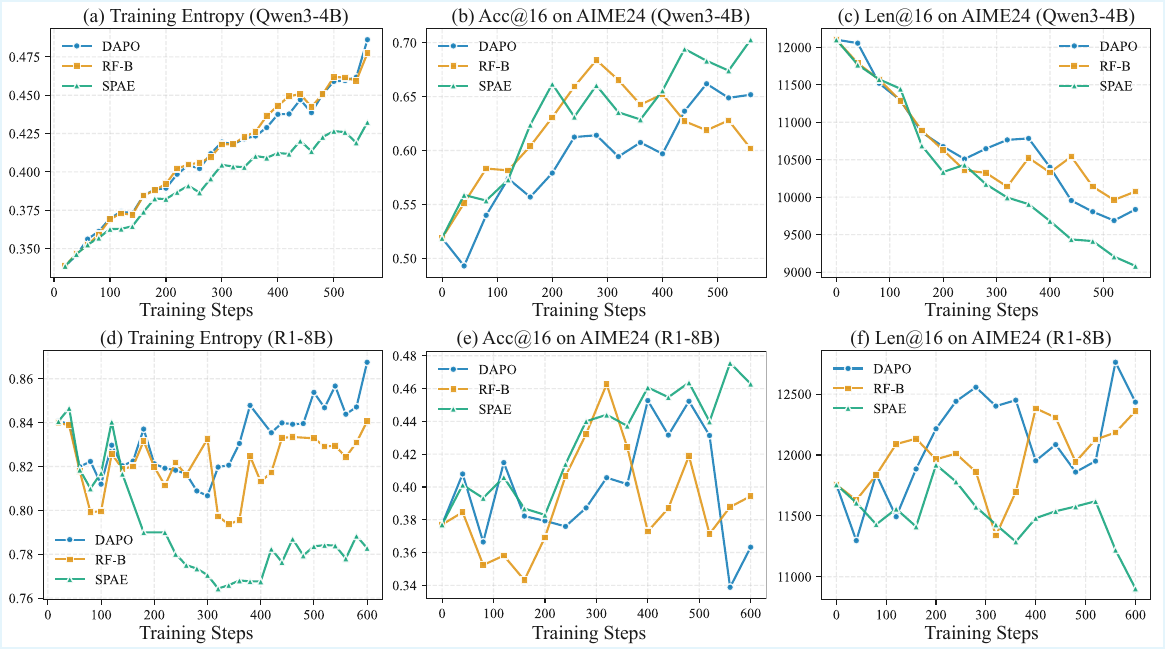}
\caption{Training dynamics on Qwen3-4B-Thinking and DeepSeek-R1-Distill-Llama-8B.}
\label{fig:dynamics_llama8b}
\end{figure*}

\subsection{Evaluation Details}

\subsubsection{Verification Protocol}
To minimize false negatives from formatting variations, we employ a hybrid verification pipeline consistent across all methods. We first apply standard rule-based extraction via \textbf{Math Verify}\footnote{\url{https://github.com/huggingface/Math-Verify}}. As a fallback for rejected answers, we utilize \textbf{xVerify-3B-Ia} \citep{chen2025xverifyefficientanswerverifier} to judge semantic equivalence with the ground truth.

\section{Extended Experimental Results}
\label{sec:ext_results}

This section provides extended experimental results that complement the main text.
We report (i) Pass@16 (i.e., at least one correct out of 16 generations) scores for all backbones, and (ii) training dynamics for additional backbones not shown in the main paper, including DeepSeek-R1-Distill-Llama-8B and Qwen3-4B-Thinking. For DeepSeek-R1-Distill-Qwen-7B training curves are presented in the main text; the curves for these three backbones exhibit highly similar trends.

\paragraph{Pass@16 Across Benchmarks.}
Table~\ref{tab:pass16_results} summarizes Pass@16 results over all the benchmarks. Overall, SPAE achieves consistently strong performance across different backbones and evaluation sets. In particular, on DeepSeek-R1-Distill-Qwen-7B, SPAE improves Pass@16 on AIME24 and yields the best (or tied-best) performance on multiple benchmarks, demonstrating that step-aware credit assignment can translate into higher sample-level success rates. On other backbones, SPAE remains competitive with strong RLVR baselines, indicating good transferability of the proposed shaping strategy.

\paragraph{Training Dynamics: Entropy, Accuracy, and Length.}
Figure~\ref{fig:dynamics_llama8b} show training curves for DeepSeek-R1-Distill-Llama-8B and Qwen3-4B-Thinking, respectively. We track (1) the average token entropy during training as a proxy for policy uncertainty, (2) Acc@16 on AIME24 as a task-level performance indicator, and (3) Len@16 on AIME24 to measure inference cost. Across both backbones, SPAE exhibits a consistent pattern: lower entropy than DAPO and RF-B, while achieving higher Acc@16 and shorter Len@16. The similarity of these curves to those reported for DeepSeek-R1-Distill-Qwen-7B in the main text suggests that the benefits of SPAE are not model-specific but reflect a stable optimization effect induced by step-level potential shaping.

\begin{figure*}[t]
\centering
\begin{tcolorbox}[
    colback=white,
    colframe=gray!70,
    arc=4pt,
    boxrule=0.8pt,
    width=0.99\linewidth, 
    fontupper=\ttfamily\scriptsize, 
    title=\textbf{Prompt for Answer Identification},
    coltitle=white,
    colbacktitle=gray!70,
]
You are an expert annotator for long-form mathematical reasoning.\\
\\
Task: Identify the earliest sentence in the model response where the correct final answer is \textbf{first derived, calculated, or established}.\\
\\
You will be given:\\
1) PROBLEM: the original question.\\
2) GOLD\_ANSWER: the verified correct final answer (canonical).\\
3) VERIFIED\_RESPONSE: a model response that has already been verified as correct overall.\\
\\
Definition:\\
- A "sentence" is a contiguous span of text in VERIFIED\_RESPONSE ending with a sentence boundary (e.g., '.', '!', '?', or a line break).\\
- The "first obtained answer sentence" is the earliest point where the reasoning is complete and the correct value is present.\\
- \textbf{Criteria for selection:}\\
\hspace*{1em} 1. \textbf{Calculation Completion:} Select the sentence where the final calculation is performed and the result equals GOLD\_ANSWER (e.g., "Thus, 10 + 5 = 15" counts if the answer is 15).\\
\hspace*{1em} 2. \textbf{Logical Equivalence:} Select the sentence that contains an expression mathematically equivalent to the GOLD\_ANSWER, provided no further steps are needed (e.g., "The value is sqrt(4)" counts if the answer is 2).\\
\hspace*{1em} 3. \textbf{Implicit Finality:} You must identify the answer \textbf{even if it is not explicitly labeled} as "The answer is..." or "Final Answer:". If the text stream has reached the correct value naturally, that sentence counts.\\
- \textbf{Exclusions:}\\
\hspace*{1em} - Do NOT choose sentences that only set up the equation (e.g., "We need to calculate 10+5") without showing the result.\\
\hspace*{1em} - Do NOT choose later restatements, summaries, or boxed answers if the correct value was already derived in a previous sentence.\\
\\
Output format (strict):\\
Return ONLY the exact sentence (verbatim) from VERIFIED\_RESPONSE.\\
Do not add quotes, explanations, line numbers, or any extra text.\\
\\
PROBLEM:\\
\{problem\}\\
\\
GOLD\_ANSWER:\\
\{gold\_answer\}\\
\\
VERIFIED\_RESPONSE:\\
\{verified\_response\}
\end{tcolorbox}
\caption{The annotation prompt used to identify the earliest sentence that can reach the ground-truth answer.}
\label{fig:annotation_prompt}
\end{figure*}

\section{Pseudocode of SPAE}
\label{app:pseudocode}
Algorithm~\ref{alg:spae} outlines the complete training pipeline of SPAE.

\begin{algorithm*}[t]
\caption{Step Potential Advantage Estimation (SPAE)}
\label{alg:spae}
\small
\SetAlgoLined
\DontPrintSemicolon
\SetKwInOut{Input}{Input}
\SetKwInOut{Output}{Output}

\Input{Dataset $\mathcal{D}$, Policy $\pi_{\theta}$, Group size $G$, Shaping weight $\xi$, Penalty strength $\alpha$}
\Output{Optimized Policy $\pi_{\theta^*}$}

\textbf{Initialize}: Policy parameters $\theta \leftarrow \theta_0$.\;

\For{each training iteration}{
    \tcp{1. Group Sampling}
    Sample a batch of queries $\mathcal{B}_q \sim \mathcal{D}$.\;
    \For{each query $q \in \mathcal{B}_q$}{
        Generate group responses $\{o_1, \dots, o_G\} \sim \pi_{\theta}(\cdot|q)$.\;
        Compute binary rewards $\{R_i=R(o_i,y^*)\}_{i=1}^G$.\;
    }

    \tcp{2. Probing \& Step Potential Computation}
    Initialize increment set $\mathcal{S}_{\Delta} \leftarrow \emptyset$.\;
    \ForEach{response $o_i$ in batch}{
        Parse reasoning steps $\tau_i = [\tau_i^{1}, \dots, \tau_i^{K_i}]$.\;
        \For{$k \leftarrow 1$ \KwTo $K_i$}{
            Construct probe context $h_{i,k} \leftarrow (q, o_{i,\le k}, p_{\text{probe}})$.\;
            Sample $N$ continuations to estimate $\mathrm{Conf}(\tau_i^{k})$ and $\mathrm{Acc}(\tau_i^{k})$.\;
            Compute Step Potential $\Phi(\tau_i^{k})$.\;

            \If{$k \ge 2$}{
                Compute increment $\Delta \Phi_{i,k} \leftarrow \Phi(\tau_i^{k}) - \Phi(\tau_i^{k-1})$.\;
                Add $\Delta \Phi_{i,k}$ to $\mathcal{S}_{\Delta}$.\;
            }
        }
    }

    \tcp{3. Advantage Estimation (Penalty \& Shaping)}
    Compute Min--Max normalization statistics from $\mathcal{S}_{\Delta}$ within the training batch $\mathcal{B}$.\;

    \ForEach{response $o_i$}{
        Compute Group Advantage $\hat{A}_i^{\text{Group}} \leftarrow R_i - \mathrm{mean}(\{R_k\}_{k=1}^G)$.\;

        \tcp{Pre-compute step-level penalty and shaping for this trajectory}
        \For{$k \leftarrow 1$ \KwTo $K_i$}{
            Compute saturation-count:
            $N_{\text{sat}}^{(i,k)} \leftarrow \sum_{t=1}^{k-1}\mathbb{I}\!\left[\Phi(\tau_i^{t}) \ge 1-\varepsilon\right]$.\;
            Compute saturation penalty:
            $f(\Phi_{i,k}) \leftarrow 1 - \alpha\left(1 - \exp\!\left(-N_{\text{sat}}^{(i,k)}\right)\right)$.\;

            \If{$k \ge 2$}{
                Let $\Delta \tilde{\Phi}_{i,k}$ be the Min--Max normalized value of $\Delta \Phi_{i,k}$ within $\mathcal{B}$.\;
                Compute shaping signal:
                $g(\Delta \Phi_{i,k}) \leftarrow \exp(\Delta \tilde{\Phi}_{i,k}) - \mathbb{E}_{(i',k')\in\mathcal{B}}\!\left[\exp(\Delta \tilde{\Phi}_{i',k'})\right]$.\;
            }
            \Else{
                Set $g(\Delta \Phi_{i,1}) \leftarrow 0$.\;
            }
        }

        \ForEach{token $j$ in $o_i$}{
            Map token $j$ to step index $k \leftarrow \mathcal{M}(j)$.\;
            \textbf{Compute SPAE advantage:}
            $\hat{A}_{i,j}^{\text{SPAE}} \leftarrow \hat{A}_i^{\text{Group}} \cdot f(\Phi_{i,k}) + \xi \cdot g(\Delta \Phi_{i,k})$.\;
        }
    }

    \tcp{4. Global Normalization \& Policy Update}
    Collect all $\hat{A}^{\text{SPAE}}$ in batch to compute mean $\mu_{\mathcal{B}}$ and std $\sigma_{\mathcal{B}}$.\;
    Normalize:
    $\hat{A}_{i,j}^{\text{Final}} \leftarrow (\hat{A}_{i,j}^{\text{SPAE}} - \mu_{\mathcal{B}}) / (\sigma_{\mathcal{B}} + \epsilon)$.\;
    Update $\theta$ by maximizing the RL objective using $\hat{A}_{i,j}^{\text{Final}}$.\;
}
\end{algorithm*}


\end{document}